\documentclass[10pt,twocolumn,letterpaper]{article}

\usepackage{iccv}
\usepackage{times}
\usepackage{epsfig}
\usepackage{graphicx}
\usepackage{amsmath}
\usepackage{amssymb}
\usepackage{float}
\usepackage{lipsum}

\usepackage[pagebackref=true,breaklinks=true,letterpaper=true,colorlinks,bookmarks=false]{hyperref}

\iccvfinalcopy 

\def\iccvPaperID{16} 

\ificcvfinal\fi

\begin{document}

\title{A Horse with no Labels: Self-Supervised Horse Pose Estimation from Unlabelled Images and Synthetic Prior}

\author{Jose Sosa, David Hogg\\
School of Computing, University of Leeds\\
{\tt\small \{scjasm, D.C.Hogg\}@leeds.ac.uk}
}

\twocolumn[{
\renewcommand\twocolumn[1][]{#1}%
\maketitle
\begin{center}
    \centering
    \includegraphics[width=\textwidth]{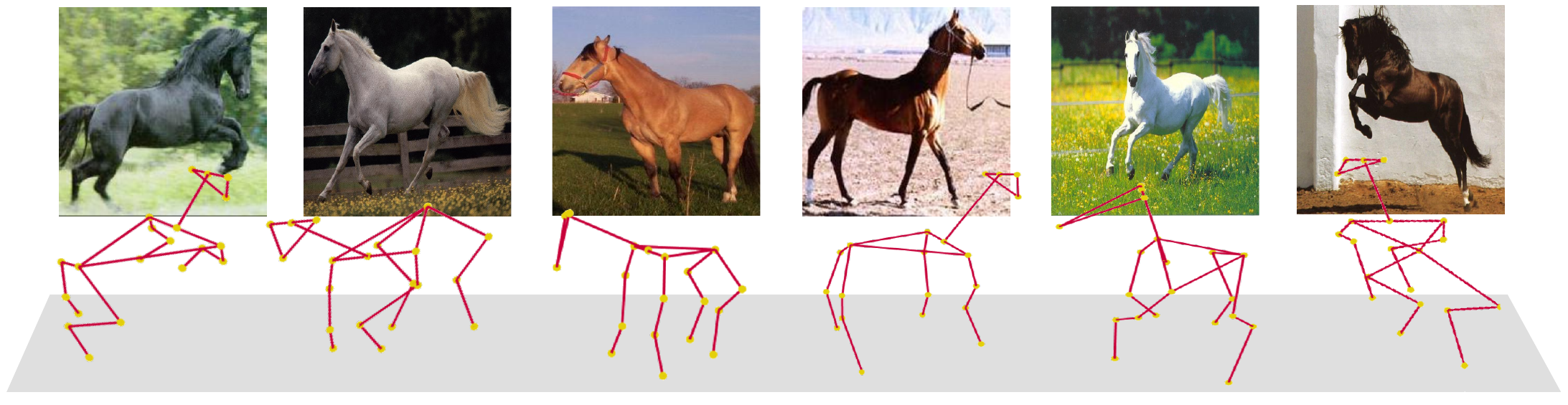}
\end{center}}]

\ificcvfinal\thispagestyle{empty}\fi


\begin{abstract}
Obtaining labelled data to train deep learning methods for estimating animal pose is challenging. Recently, synthetic data has been widely used for pose estimation tasks, but most methods still rely on supervised learning paradigms utilising synthetic images and labels. Can training be fully unsupervised? Is a tiny synthetic dataset sufficient? What are the minimum assumptions that we could make for estimating animal pose? Our proposal addresses these questions through a simple yet effective self-supervised method that only assumes the availability of unlabelled images and a small set of synthetic 2D poses. We completely remove the need for any 3D or 2D pose annotations (or complex 3D animal models), and surprisingly our approach can still learn accurate 3D and 2D poses simultaneously. We train our method with unlabelled images of horses mainly collected for YouTube videos and a prior consisting of 2D synthetic poses. The latter is three times smaller than the number of images needed for training. We test our method on a challenging set of horse images and evaluate the predicted 3D and 2D poses. We demonstrate that it is possible to learn accurate animal poses even with as few assumptions as unlabelled images and a small set of 2D poses generated from synthetic data. Given the minimum requirements and the abundance of unlabelled data, our method could be easily deployed to different animals.

\end{abstract}

\section{Introduction}

One of the main bottlenecks for supervised animal pose estimation is obtaining pose annotations for training deep-learning models. While plenty of labelled data is available in the human domain, annotated animal datasets are scarce. In order to overcome the annotation issue, new alternatives have been adopted, such as training models with synthetic data. 


We adapt a method from the human domain that learns human 3D poses from unlabelled images and a prior on 2D poses \cite{sosa2023self}. Our implementation translates this method to the animal domain, demonstrating that it applies to different body structures. Another essential addition to our approach is the origin of the 2D poses composing the prior. Unlike the original implementation \cite{sosa2023self}, which uses a set of unpaired 2D poses from the training datasets, we further reduce the assumptions by using 2D poses from an existing CAD model of a horse \cite{mu2020learning}. Our model is unique in its simplicity compared with previous approaches for animal pose estimation with synthetic data. It does not require annotated training data. It uses only unlabelled images and a small set of synthetically generated 2D poses, which means that no synthetic images, pre-trained models, or complicated 3D models are required.

We train and test the model with unlabelled images of horses. Additionally, we use a prior on 2D pose generated from synthetic data \cite{mu2020learning}. By evaluating 2D and 3D predictions from our model, we demonstrate that our approach produces accurate 2D and 3D pose representations of the horses, although we are not using any annotations for the input images. Since the requirements for training our model are minimal, it could be easily applied to animal species with different body structures.

\section{Related work}
\subsection{Animal pose estimation}

Supervised deep learning methods for human pose estimation have been widely explored and perform well under different conditions \cite{newell2016stacked,toshev2014deeppose,martinez2017simple}. However, in animal pose estimation, getting the labels needed for supervision is difficult in most cases. In particular, labelling key points is more expensive and laborious than producing other annotations, e.g. bounding boxes. On top of this, it would be infeasible to generate labelled data for the entire diversity of animal species in the world.

Since the 3D pose annotations are even more challenging to acquire than the 2D ones, many works on animal pose estimation have been focused only on estimating 2D pose \cite{mathis2021pretraining,mathis2018deeplabcut,pereira2019fast,russello2022t}. Not surprisingly, the backbones for most of these approaches are network architectures initially designed for the human domain, for example, stacked hourglass networks \cite{newell2016stacked}, ResNet \cite{he2016deep}, and OpenPose\cite{cao2017realtime}.

Although the problem of 3D animal pose estimation is more constrained and challenging, relevant work has also been carried out \cite{bala2020automated,9561338,gosztolai2021liftpose3d}. In this context, methods commonly rely on lower supervision levels to overcome the scarcity of labelled training data. For instance, the self-supervised approach of \cite{zhang2020multiview} estimates 3D pose for monkeys and dogs relying on multi-view supervision and a tiny portion of pose annotations. Dai \etal \cite{dai2023unsupervised} proposes an similar method, but instead of multiview images, they assume the availability of actual 2D poses for each input image and lift these to 3D through self-supervision based on geometric consistency. Similar to \cite{dai2023unsupervised} our method also estimates 3D pose using self-supervision with the same geometric consistency constraint. However, we learn the 2D and 3D poses directly from images in an end-to-end manner. Most importantly we do not require any annotations for the inputs.

\subsection{Animal pose estimation with synthetic data}
Synthetic data is a low-cost alternative to generate data with ground truth annotations with minimum effort. Recently, works on human \cite{loper2015smpl,doersch2019sim2real,varol2017learning} and animal pose \cite{bolanos2021three,mu2020learning,li2021synthetic,cao2017realtime,biggs2019creatures,biggs2020left,ruegg2023bite,zuffi2018lions,zuffi2019three,sosa2022mice} estimation have adopted synthetic data to overcome the scarcity of keypoint labels.

Many animal pose estimation methods with synthetic data follow a supervised approach, meaning they use synthetically generated images and pose annotations for training. However, there is often a gap between synthetic and real data, so these approaches typically perform domain adaptation with samples from actual data. For example, \cite{mu2020learning} learns 2D pose for animals using images and labels generated from CAD models. They also incorporate a consistency-constrained semi-supervised method to adapt the predictions to real data. Similarly, \cite{li2021synthetic} focuses on domain adaptation by generating pseudo-labels from the synthetic domain and then updating these to match the actual data. Unlike these approaches, our formulation helps to reduce the complexity and requirements for training even more. It is as simple as using unlabelled real images and a set of synthetically generated 2D poses, i.e. there is no need to generate pictures from the synthetic data. Furthermore, an adversarial loss helps to learn poses that do not necessarily appear in the prior of synthetic 2D poses without having additional processes to align domains. 

More related to our work, \cite{sosa2022mice} relies on a self-supervised method that assumes synthetic 2D poses and real images for estimating 2D mouse pose. However, we advance \cite{sosa2022mice} by incorporating geometry consistency, allowing our model to further estimate 3D pose.

Synthetic data also plays an essential role in several works that learn richer structures, such as animal shapes, mainly for different quadrupeds like dogs \cite{biggs2020left,ruegg2023bite, biggs2019creatures}, tigers, lions, horses \cite{zuffi2018lions}, and zebras \cite{zuffi2019three}. However, the success of these approaches is constrained by having access to sophisticated and expensive animal models, which is not required in our approach. 

\section{Method}

\begin{figure*}
    \centering
    \includegraphics[width=\textwidth]{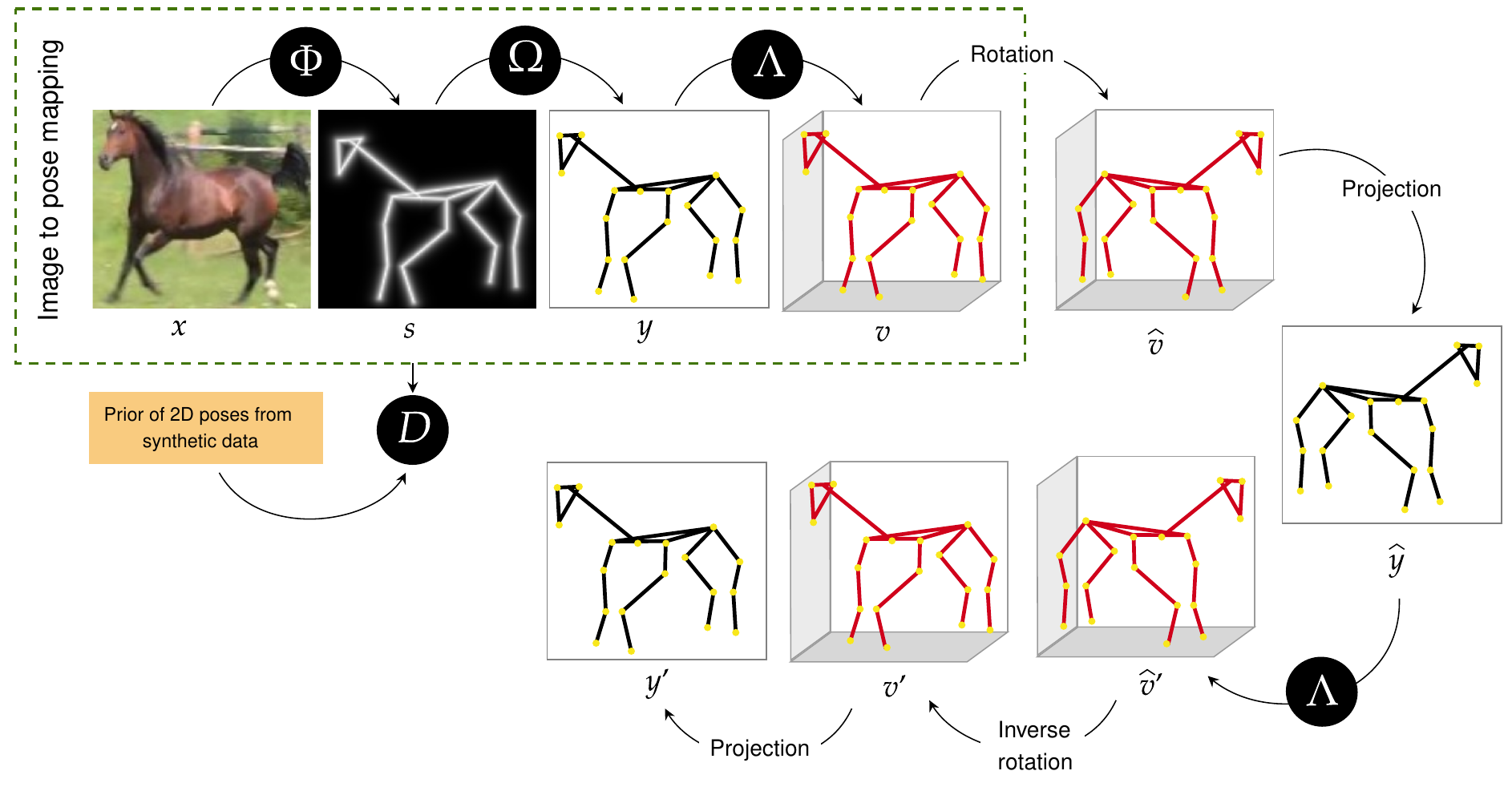}
    \caption{The method is an adaptation of \cite{sosa2023self}. The main difference is the origin of the prior on 2D poses. In this case we use a publicity available set of 2D poses from an existing CAD model of a horse from \cite{mu2020learning}.}
    \label{fig:main-model}
\end{figure*}

The method is essentially that from \cite{sosa2023self}. Unlike Sosa \cite{sosa2023self} we translate this method to the animal domain and most importantly we change the origin of the prior. In the original paper they use a prior of 2D poses coming from unpaired annotations of the training dataset. We remove this by generating the 2D poses from synthetic data. We reproduce the method here so that the current paper is self-contained.

The main component of the approach is an image to 3D pose mapping, indicated with a dotted box in \autoref{fig:main-model}. The first part of this mapping employs a CNN $\Phi$ to map the input image $x$ to an intermediate skeleton image $s$. Then, another CNN $\Omega$ maps $s$ to a 2D pose representation $y$. In the final stage, $y$ is mapped to the 3D pose $v$ by means of a fully connected network $\Lambda$. For training this set of networks, we incorporate it within a larger structure which allows for self supervision. In particular, we rely in a loop of transformations of the 3D pose $v$. We also use a discriminator $D$ together with the prior on synthetic 2D poses, to ensure that the generated skeletons $s$ are realistic. 

\subsection{Main mapping}

The image to pose mapping consists of 3 networks $\Phi$, $\Omega$, and $\Lambda$ that allows the input image $x$ to be mapped to its 3D pose representation $v$. This mapping also produces two intermediate representations of the input, a skeleton image $s$, and a 2D pose $y$. Specifically, $\Phi$ learns to align the input image with its respective skeleton image representation, i.e. $s = \Phi(x)$. Then, $\Omega$ learns to extract keypoints from $s$, obtaining a 2D pose as output $y = \Omega( \Phi (x) ) $. Finally, $\Lambda$ acts as a lifter of the 2D pose $y$ to get the 3D pose $v$. For each pair of joint positions $(x_i, y_i)$ in $y$, the network estimates a depth $z_i = d + \Delta$, where $\Delta$ is a constant depth.

Overall, we use the same network structure as in \cite{sosa2023self} with exception of $\Lambda$. Since we are not trying to learn elevation angles for the geometry transformations like \cite{sosa2023self,wandt2022elepose}, we opt for a simpler structure as in \cite{chen2019unsupervised,martinez2017simple}. 

\subsection{Self-supervision}

As illustrated by \autoref{fig:main-model}, we include the main mapping within a large network structure that allows to self-supervise the training. This structure uses a discriminator network $D$, which relies on a prior of synthetic 2D poses to help the mapping produce skeleton images that are as realistic as possible. Furthermore, it incorporates a loop of random rotations and projections of the 3D pose $v$ to ensure geometry consistency for the 3D predictions.

\subsubsection{Synthetic pose prior}
\begin{figure}[!h]
    \centering
    \includegraphics[width=\columnwidth]{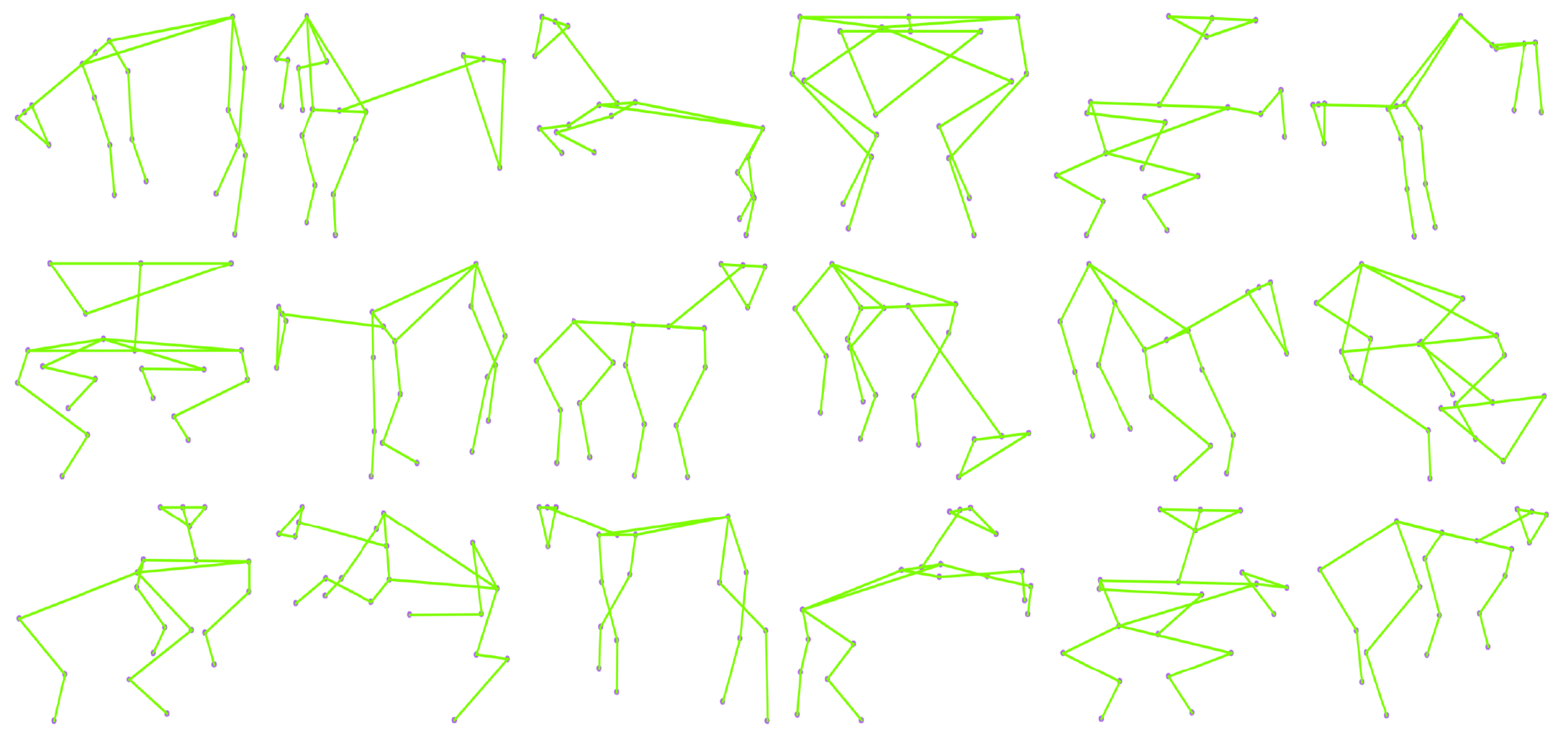}
    \caption{Random 2D poses from the syntethic prior.}
    \label{fig:prior_example}
\end{figure}

To create the prior of 2D poses, we use a publicly available dataset of synthetic 2D poses generated from a CAD model of a horse \cite{mu2020learning}. The prior is needed during training to ensure the estimated skeleton image looks as realistic as possible. Note that generating the prior from synthetic data and not from annotations of the dataset like \cite{sosa2023self} provides more flexibility to the method to be trained with completely unlabelled datasets, which are abundant in the animal domain. Our synthetic prior contains around 10k different 2D poses, representing approximately one-third of the available images for training. \autoref{fig:prior_example} provides examples of some 2D poses $p$ in the prior.

The purpose of having a prior of 2D poses is to use these as a reference distribution for the discriminator network $D$. Since our implementation of $D$ works directly with images, we must first render the synthetic 2D poses to skeleton images. This is done by using the rendering function $\beta$ from \cite{jakab2020self}, which given a set of 2D joint positions $p$ and their connections, can generate a skeleton image $w = \beta(p)$. Then, the goal of $D$ is to evaluate whether or not the predicted skeleton image $s = \Phi(x)$, looks like an authentic skeleton image $w$ such as those in the prior. Following \cite{sosa2023self,jakab2020self} we use an adversarial loss to compare $w$ and $s$:

\begin{equation}
     L_D = \mathbb{E}_w(log(D(w)) + \mathbb{E}_s(log(1 - D(s))
     \label{eq:loss_d}
\end{equation}

\subsubsection{Geometry consistency}
We rely on the idea of geometric consistency from \cite{chen2019unsupervised} to facilitate the learning of the lifting network $\Lambda$ and, therefore, the whole mapping. Essentially this involves a series of rotations and projections of the 3D pose $v$. First, $v$ is randomly rotated to $\hat{v}$ using a rotation matrix, which is constructed by sampling azimuth and elevation angles from a fixed uniform distribution \cite{chen2019unsupervised}. Then, $\hat{v}$ is projected to a 2D pose $\hat{y}$. Given the projection of the rotated 3D pose $\hat{v}$, the same lifting network $\Lambda$ estimates its 3D representation $\hat{v}'$. Lastly, the inverse rotation is applied to the 3D pose $\hat{v}'$ to obtain $v'$, and $v'$ is projected to 2D to get the 2D pose $y'$. 

After the loop of projections and rotations we expect the poses on the forward and backward parts to be as similar as possible. For example, the 3D poses $v$ and $v'$ should be similar, and the same with $\hat{v}$ and $\hat{v}'$. This also applies to the 2D poses $y$ and $y'$. Therefore, we can derive the following component loss functions:

\begin{equation}
    L_{2D} = || y' - y || ^ 2
    \label{eq:2d_loss}
\end{equation}

\begin{equation}
    L_{3D} = || (v'^{(j)} - v'^{(k)} ) - (v^{(j)} - v^{(k)} )|| ^ 2
    \label{eq:3d_loss}
\end{equation}

\begin{equation}
    L_{r3D} =  || \hat{v}' - \hat{v}|| ^ 2
    \label{eq:3d_loss_r}
\end{equation}

Note that for \autoref{eq:3d_loss} we follow \cite{wandt2022elepose, sosa2023self} and instead of comparing the $v$ and $v'$ with a $L_2$ loss we measure the degree of deformation between 3D poses using two samples $j$ and $k$ in a batch. For simplicity, we refer to the sum of these three losses as $L_{GC}$ given by 

\begin{equation}
    L_{GC} = L_{2D} +  L_{3D} + L_{r3D}
    \label{eq:loss_pose}
\end{equation}

\subsubsection{Training and additional losses}

Following \cite{jakab2020self} we include an extra loss term $L_{\Omega}$ to evaluate the mapping $y = \Omega(s)$, i.e. from the skeleton image $s$ to the 2D pose $y$.

\begin{equation}
    L_{\Omega} = || ( \Omega(\beta(p)) - p   )||^2 + \lambda|| \beta(y) - s ||^2 
    \label{eq:loss_mapping}
\end{equation}

where $\lambda$ represents a balancing coefficient, $p$ is a 2D pose from the unpaired prior and $\beta$ is rendered from \cite{jakab2020self}.

We train all the networks from scratch using a loss function $L$ consisting of three components from \autoref{eq:loss_d}, \autoref{eq:loss_pose}, and \autoref{eq:loss_mapping}.

\begin{equation}
    L = L_D + L_{GC} + L_{\Omega}
\end{equation}

At inference time, we only keep the elements from the main mapping as illustrated in the dotted box from \autoref{fig:main-model}, i.e. the loop of rotations and projections, and $D$ are only needed during training.

\section{Experiments}

\subsection{Data}
We train the model with a dataset of video frames depicting full-body horses. First, we select the horse subset from the latest version of the TigDog dataset \cite{del2017behavior}. We use video frames for all the horse sequences in the dataset, discarding video frames showing partially visible horses. To increase the diversity of horses in the training set, we automatically collect video frames for a manually defined group of YouTube videos that are expected to show horses throughout (i.e, in most video frames). To gather the video frames automatically from a video, we follow a three-step process:

\begin{enumerate}
    \item Download the video from YouTube and split it into frames.
    \item Process each frame using a pre-trained model from \cite{wu2019detectron2}, which identifies the horse and produces a segmentation mask. Remove frames that do not contain a horse. 
    \item Resize the frames showing a horse to a predefined size ($128 \times 128$) and save them along with their respective segmentation mask generated by the model. 
\end{enumerate}

We collect frames containing complete horses from about 60 videos, representing 47k frames (plus around 6k from \cite{del2017behavior}). Note that this dataset is relatively small compared to what is required for training human pose estimation models — our horse dataset is only 1.3\% of the size of the Human3.6M dataset \cite{ionescu2013human3} and 3.6\% of the size of the MPI-INF-3DHP dataset \cite{mono-3dhp2017}.

\begin{figure*}[h]
    \centering
    \includegraphics[width=\textwidth]{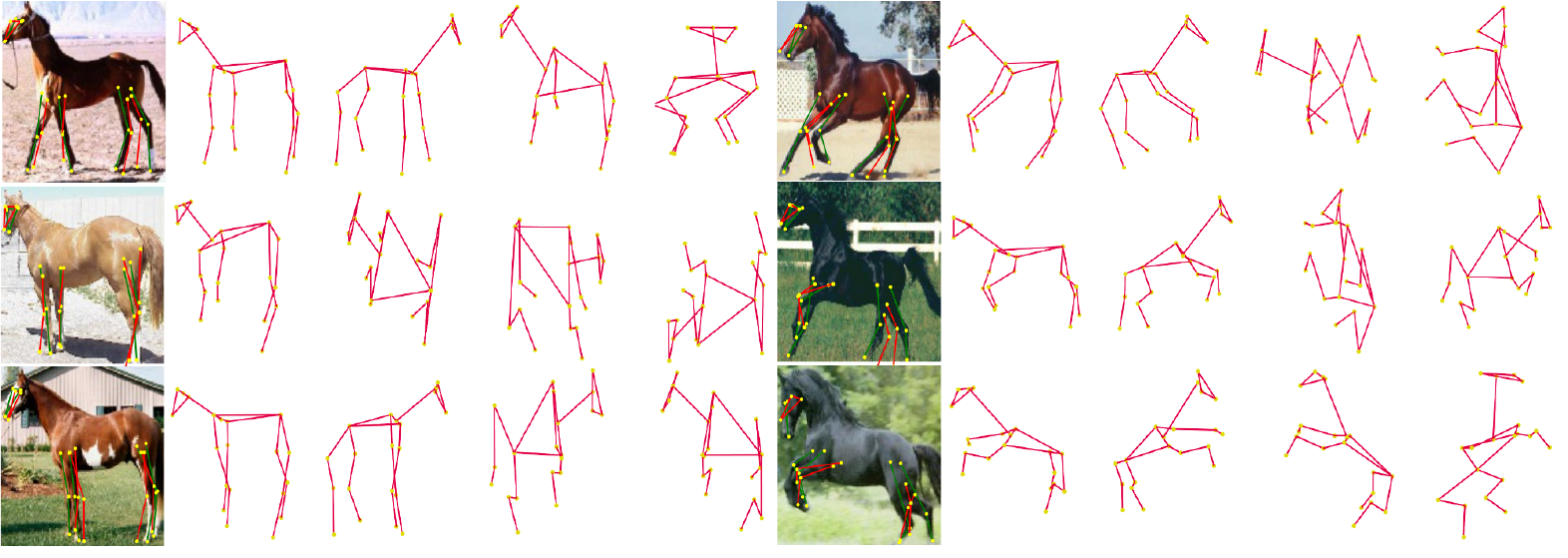}
    \caption{ \textbf{3D poses estimated by our method.}  The first and sixth columns show the images with their estimated (red) and ground truth (green) 2D poses. The rest of the columns illustrate novel views of the predicted 3D pose.}
    \label{fig:3d-poses}
\end{figure*}

\subsubsection{Test data}
In real life applications we cannot assume that test data will come from the same source as the training data. Thus, instead of selecting a hold-out set of frames for each training video (which potentially could lead to better performance), we use more challenging data to test our model. In particular, we utilise images from a different collection: the Weizmann dataset \cite{borenstein2004combining}. However, this data does not contain annotations for 2D or 3D poses. We therefore manually annotate the 2D poses consisting of 15 joint positions (3 for each front and rear limb, 1 for the chin, and 2 for the eyes) for all the images in the Weizmann dataset showing full-body horses (around 300). We use the pose annotations as ground truth to quantitatively evaluate the estimated 2D poses from our method.

\subsection{Evaluation}
Since there is a lack of available horse datasets with 3D pose annotations for a quantitative performance evaluation, we only assess the quality of the 3D predictions qualitatively. While obtaining 2D poses is more feasible than 3D poses, we evaluate the emergent 2D pose predictions $y$ quantitatively and qualitatively. Note that although the goal of the model is predicting 3D poses, the emergent 2D pose representations are also worth evaluating. We assume that if the 2D poses are good, it is very likely the 3D poses will be reasonable as well.

In line with previous works for 2D animal pose estimation, we use the Percentage of Correct Keypoints (PCK@0.05) to quantitatively evaluate our 2D predictions. Our predicted poses are composed of 20 joint positions. However, we use only 15  in order to compare with the ground truth 2D poses from the Weizmann dataset.

\subsection{Results \& Discussion}
\subsubsection{Results on 3D pose predictions}
Given the scarcity of ground truth 3D data for horses, we provide only a qualitative evaluation of the 3D poses estimated by our trained model in \autoref{fig:3d-poses}.

\begin{figure}
    \centering
    \includegraphics[width=\columnwidth]{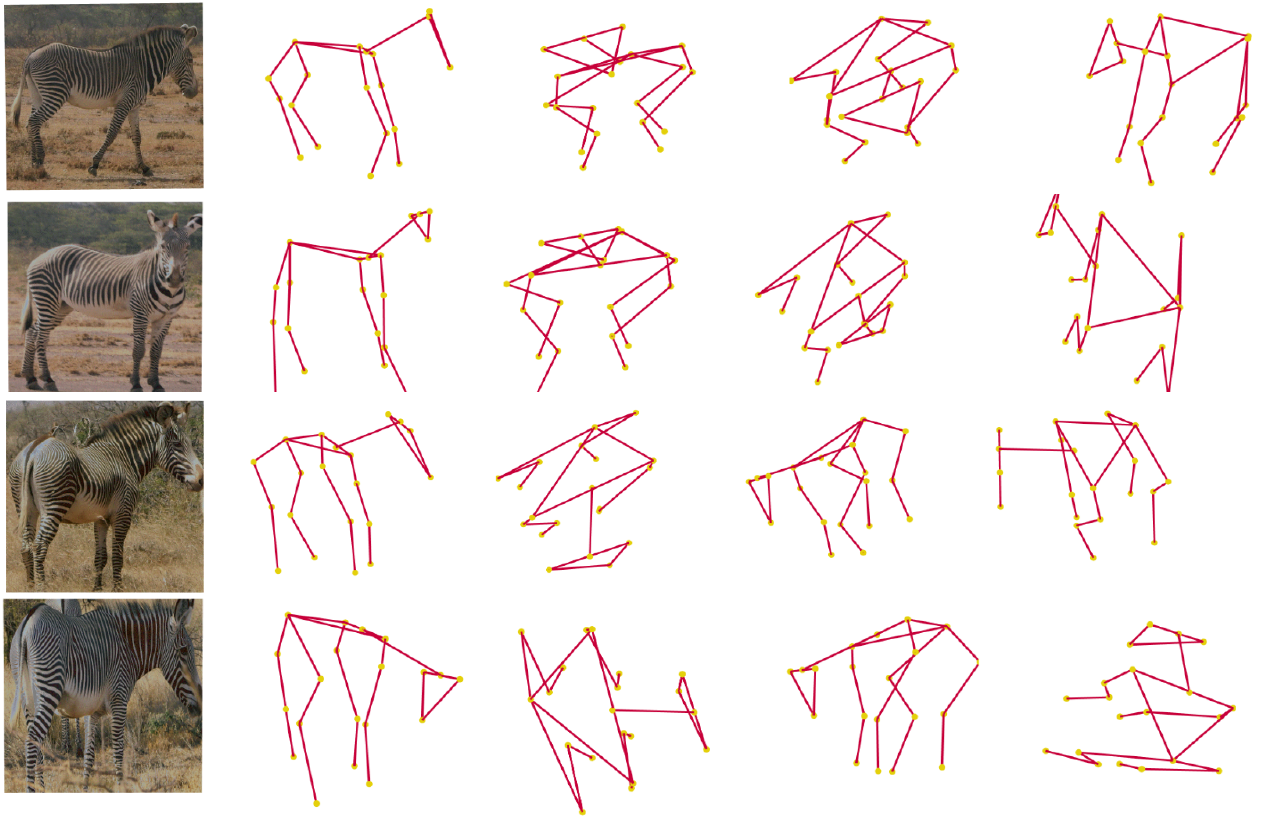}
    \caption{\textbf{3D pose predictions for zebras.} First column shows the input image. Following columns show novel views of the 3D poses.}
    \label{fig:zebras-results}
\end{figure}

Additionally, we test the generalisation capability of our model by evaluating it on a dataset of zebras \cite{zuffi2019three}. Because of the anatomical similarities between zebras and horses, the trained model with the horse data can still estimate plausible 3D poses for zebras (although it has never seen a zebra during training). \autoref{fig:zebras-results} displays some 3D predictions for zebras. Given the slight differences between the two species (zebras having slightly wider chests and shorter legs), these results show the robustness of our method to different domains.

\begin{table*}[ht]
  \centering
  \setlength\tabcolsep{0pt}
  \begin{tabular*}{\textwidth}{@{\extracolsep{\fill}} llcccccc }
    \hline
    Method & Evaluation Data &  Eyes & Chin & Shoulders & Knees & Hooves & \textbf{Mean} \\
    \hline
    
    Syn - Mu \etal \cite{mu2020learning}&  TigDog dataset & 46.08  & 53.86 & 20.46  &   24.20 &   17.45   & 25.33\\
    
    Sosa \cite{sosa2022mice} & Weizmann dataset & 45.67 & 44.67 & 33.00 & 37.67 & 26.67 & 37.54\\
    CycleGAN \cite{zhu2017unpaired} & TigDog dataset & 70.73   & 84.46 & 56.97 &  49.91 &   35.95 &    51.86\\
    \hline
    
    Ours & Weizmann dataset & 49.3 & 58.3 & 34.2 & 44.7 & 31.2  & 43.50 \\
    \hline
  \end{tabular*}
  \caption{\textbf{Horse 2D pose estimation accuracy.} We calculate the accuracy of our predicted 2D poses using the PCK@0.05 metric. For each image in the Weizmann dataset, the predicted 2D pose is compare against its respective ground truth. We also list some works that estimate 2D poses using synthetic data.}
  \label{tab:quantitative-comp} 
\end{table*}

\subsubsection{Results on 2D pose predictions}

\begin{figure} [ht]
    \centering
    \includegraphics[width=\columnwidth]{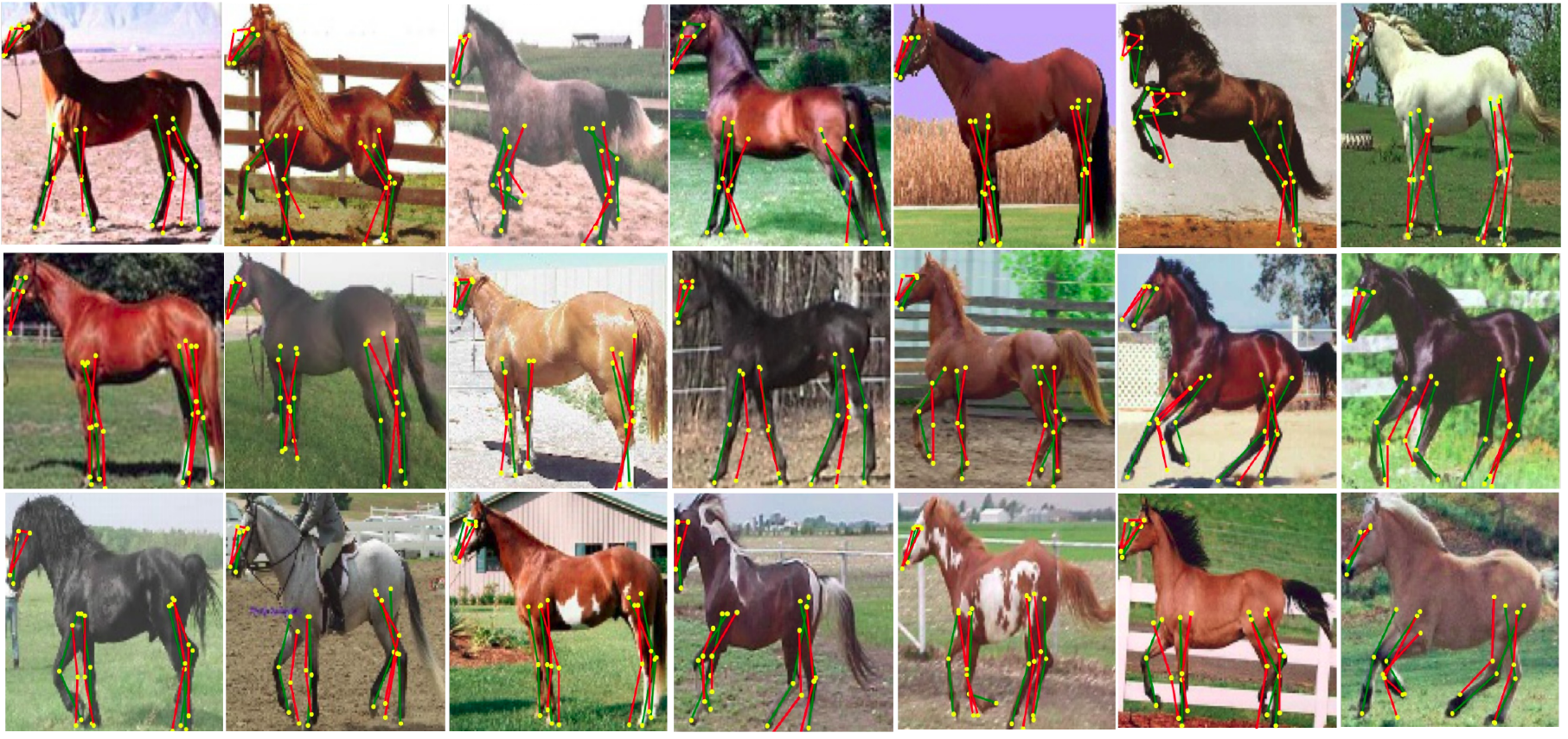}
    \caption{\textbf{Predicted 2D poses with our method.} Each image comes from the test set; the red lines represent the connections between our method's estimated 2D joint positions. The green lines represent the connections between the ground truth joint positions.}
    \label{fig:2d-poses}
\end{figure}

Using the trained model, we produce 2D poses for all the images in the test set. Each pose prediction consists of 20 joint positions. However, when comparing against ground truth, we only keep 15 joint positions to match the annotations. \autoref{fig:2d-poses} shows some predicted 2D poses by our model compared with their respective ground truth. 

\begin{figure}
    \centering
    \includegraphics[width=\columnwidth]{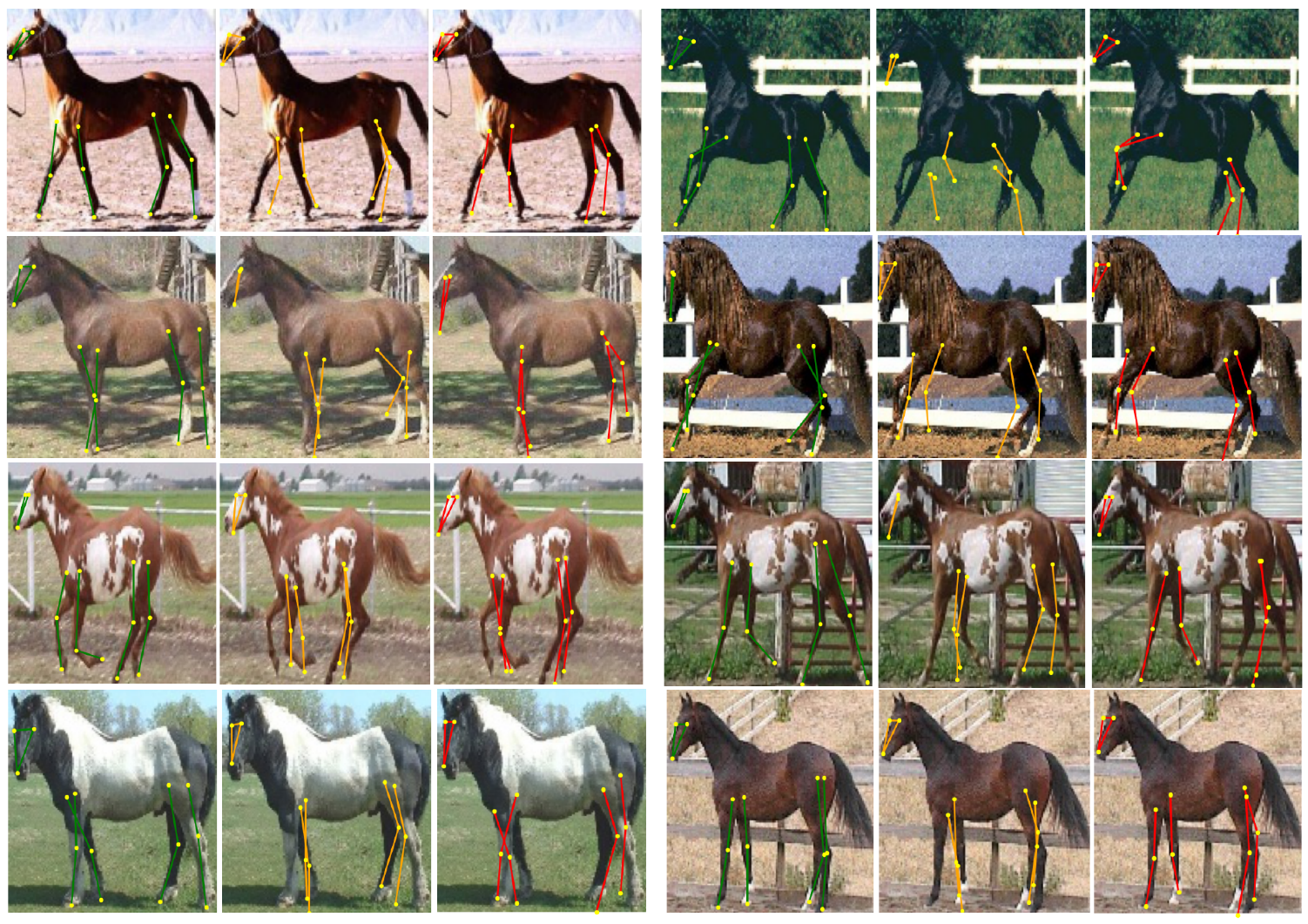}
    \caption{\textbf{Comparison of 2D pose predictions with a similar method.} The first and fourth columns show the ground truth joint positions (green). The second and fifth columns show the estimated 2D poses by \cite{sosa2022mice} (orange). The third and sixth columns display the estimated 2D poses by our method (red).}
    \label{fig:comparison-2d}
\end{figure}

In addition, we reproduce the method from \cite{sosa2022mice} that originally estimates 2D poses for mice. We train it with the same assumptions that our method, i.e. our same horse dataset and synthetic 2D poses. We use the Weizmann dataset to evaluate and compare their predictions with the ones obtained with our 3D method. As illustrated by \autoref{fig:comparison-2d}, our model for 3D poses can produce more accurate 2D pose representations than the 2D pose estimator from \cite{sosa2022mice}. This comparison demonstrates the value of incorporating the geometry consistency idea for lifting 2D poses to 3D. 

We use the PCK@0.05 metric to evaluate the predicted 2D poses against their respective ground truth. \autoref{tab:quantitative-comp} shows the accuracy results of our quantitative evaluation for 2D pose. It also includes results for approaches that work under similar conditions. However, note that except for \cite{sosa2022mice}, which assumes the same setting as our method, the others methods apply supervised learning during training. Although the performance is not better than some of the methods listed in the table, it is also competitive, given the minor requirements of our method.

\begin{figure}
    \centering
    \includegraphics[width=\columnwidth]{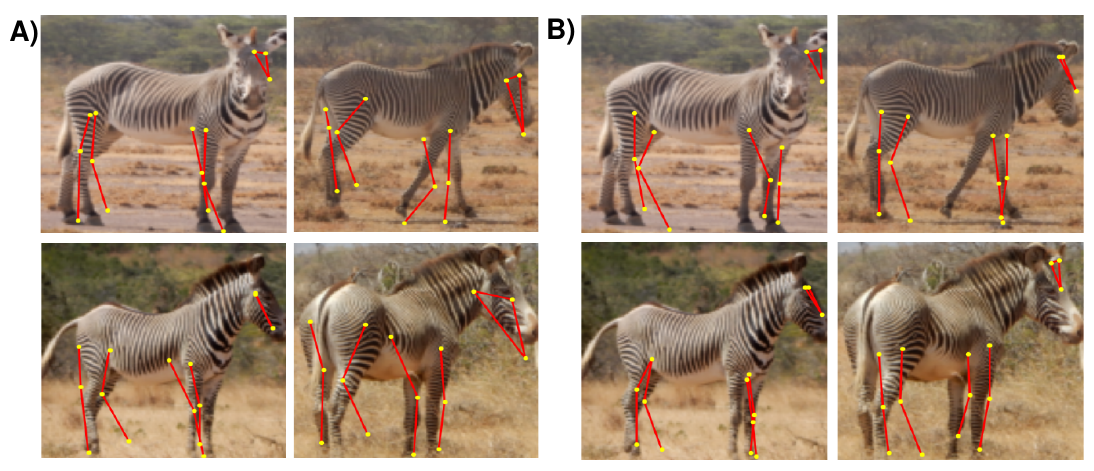}
    \caption{\textbf{Predicted 2D poses.} Block A, shows 2D poses predicted by the model trained with images of horses and the prior on synthetic 2D poses. Block B, display 2D predictions using synthetic images of zebras and a the same prior on synthetic 2D poses from horses.}
    \label{fig:zebras-results-2d}
\end{figure}

Furthermore, we experiment by training our method on the synthetically generated images of zebras from \cite{zuffi2019three}, and using the same synthetic 2D horse poses as prior. 
We then test on the same dataset of real zebras \cite{zuffi2019three} as in previous experiments (model trained with horse images and synthetic 2D poses as prior). Despite the differences between domains, the model trained with purely synthetic data (synthetic images of zebras and synthetic 2D poses of horses) produces similar 2D poses as the model trained with real horse images and the synthetic 2D horse poses. \autoref{fig:zebras-results-2d} shows the predicted 2D poses for different images from both configurations.

\subsubsection{Failed cases}

\begin{figure}[ht]
    \centering
    \includegraphics[width=\columnwidth]{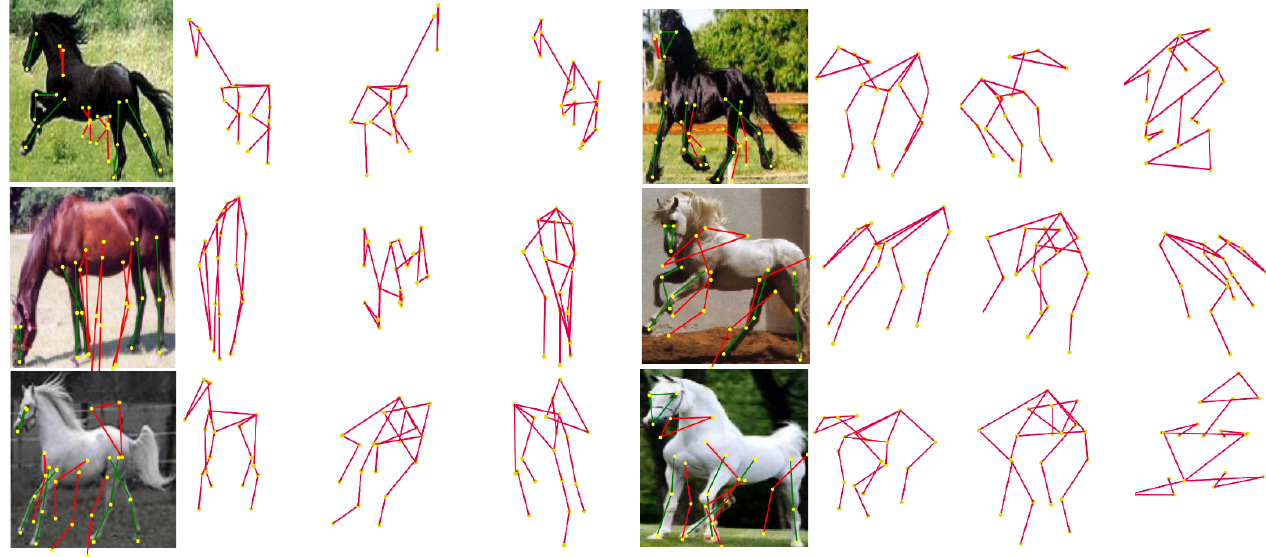}
    \caption{\textbf{Failed cases for the estimated 2D poses.} We select the 2D poses with lower accuracy (PCK@0.05) to have a look at their respective 3D predictions. }
    \label{fig:failed-2d}
\end{figure}

Note that the quality of the emergent 2D pose estimations influences the accuracy of the final 3D pose predictions. Therefore, when the previous 2D predictions depict proper horse poses, the 3D predictions are expected to be more accurate. Surprisingly, even for some non-accurate 2D predictions, our model can still recover a plausible 3D horse pose, as shown in \autoref{fig:failed-2d}.

\section{Conclusion}

We have successfully adapted a method originally designed to estimate 3D human poses to the animal domain. We further reduce its requirements by generating the needed prior from synthetic data. We show that with only unlabelled images and a small set of synthetic 2D poses, it is possible to learn 3D representations. By reducing the data requirements for training to a minimum, our proposal could be applied to many unlabelled detests without collecting annotations needed for supervised training. 

From our results, there is clearly room for further improvement. Two ideas for exploration in future are (1) to incorporate temporal information into the approach, and (2) to follow previous work in fine-tuning with small amounts of actual data to reduce the gap between the synthetic and real domains.

\section*{Acknowledgments}
Thanks to Rebecca Stone from the School of Computing at University of Leeds for useful feedback and great discussions.

{\small
\bibliographystyle{ieee_fullname}
\bibliography{egbib}
}

\end{document}